\documentclass[12pt]{article}

\usepackage{amsmath,amsthm, amsfonts, amssymb, amsxtra,amsopn}
\usepackage{pgfplots}
\usepgfplotslibrary{colormaps}
\pgfplotsset{compat=1.15}
\usepackage{pgfplotstable}
\usetikzlibrary{pgfplots.statistics} 
\usepackage{filecontents}
\usepackage{graphicx}
\usepackage{multirow}
\usepackage{tabu}
\usepackage{algorithmic}
\usepackage{longtable}
\usepackage{booktabs}
\usepackage{listings}
\usepackage{cmap}
\usepackage{colortbl}
\usepackage{adjustbox}
\usepackage{epsfig}
\usepackage{xstring}%%% needed for \rowvec macros

\usepackage{subfigure}

\PassOptionsToPackage{hyphens}{url}

\usepackage{hyperref}
\hypersetup{colorlinks=true,linkcolor=black,citecolor=black,urlcolor=blue,filecolor=black}
\hypersetup{pdfpagemode=UseNone,pdfstartview=}

\usepackage[tableposition=top,width=1.0\textwidth]{caption}
\usepackage{enumitem}
\setlist[itemize]{noitemsep, topsep=0pt}

\advance\oddsidemargin by -0.45in
\advance\textwidth by 0.9in

\advance\topmargin by -0.3in
\advance\textheight by 0.6in

\long\def\symbolfootnotetext[#1]#2{\begingroup%
\def\thefootnote{\fnsymbol{footnote}}\footnotetext[#1]{#2}\endgroup}

\pgfkeys{
    /pgf/number format/fixed zerofill=true }
    
\pgfplotstableset{
    color cells/.code={%
        \pgfqkeys{/color cells}{#1}%
        \pgfkeysalso{%
            postproc cell content/.code={%
                \begingroup
                \pgfkeysgetvalue{/pgfplots/table/@preprocessed cell content}\value
\ifx\value\empty
\endgroup
\else
                \pgfmathfloatparsenumber{\value}%
                \pgfmathfloattofixed{\pgfmathresult}%
                \let\value=\pgfmathresult
                \pgfplotscolormapaccess
                    [\pgfkeysvalueof{/color cells/min}:\pgfkeysvalueof{/color cells/max}]%
                    {\value}%
                    {\pgfkeysvalueof{/pgfplots/colormap name}}%
                \pgfkeysgetvalue{/pgfplots/table/@cell content}\typesetvalue
                \pgfkeysgetvalue{/color cells/textcolor}\textcolorvalue
                \toks0=\expandafter{\typesetvalue}%
                \xdef\temp{%
                    \noexpand\pgfkeysalso{%
                        @cell content={%
                            \noexpand\cellcolor[rgb]{\pgfmathresult}%
                            \noexpand\definecolor{mapped color}{rgb}{\pgfmathresult}%
                            \ifx\textcolorvalue\empty
                            \else
                                \noexpand\color{\textcolorvalue}%
                            \fi
                            \the\toks0 %
                        }%
                    }%
                }%
                \endgroup
                \temp
\fi
            }%
        }%
    }
}

\def\srowvecc#1#2{(\!\begin{array}{cc} 
      \noexpandarg\IfBeginWith{#1}{-}{\! #1}{#1}
    & #2\kern-0.5pt\end{array}\!)}
\def\rowvecc#1#2{\left(\!\begin{array}{cc} 
      \noexpandarg\IfBeginWith{#1}{-}{\! #1}{#1}
    & #2\kern-0.5pt\end{array}\!\right)}
\def\rowveccc#1#2#3{\left(\!\begin{array}{ccc} 
      \noexpandarg\IfBeginWith{#1}{-}{\! #1}{#1}
    & #2 
    & #3\kern-0.5pt\end{array}\!\right)}
\def\rowvecccc#1#2#3#4{\left(\!\begin{array}{cccc}
      \noexpandarg\IfBeginWith{#1}{-}{\! #1}{#1}
    & #2 
    & #3 
    & #4\kern-0.5pt\end{array}\!\right)}
\def\srowvecccc#1#2#3#4{\bigl(\!\begin{array}{cccc}
      \noexpandarg\IfBeginWith{#1}{-}{\! #1}{#1}
    & #2 
    & #3 
    & #4\kern-0.5pt\end{array}\!\bigr)}
\def\rowveccccc#1#2#3#4#5{\left(\!\begin{array}{ccccc} 
      \noexpandarg\IfBeginWith{#1}{-}{\! #1}{#1}
    & #2
    & #3
    & #4
    & #5\kern-0.5pt\end{array}\!\right)}
\def\srowvecccccc#1#2#3#4#5#6{(\!\begin{array}{cccccc} 
      \noexpandarg\IfBeginWith{#1}{-}{\! #1}{#1}
    & #2
    & #3
    & #4
    & #5
    & #6\kern-0.5pt\end{array}\!)}
\def\rowvecccccc#1#2#3#4#5#6{\left(\!\begin{array}{cccccc} 
      \noexpandarg\IfBeginWith{#1}{-}{\! #1}{#1}
    & #2
    & #3
    & #4
    & #5
    & #6\kern-0.5pt\end{array}\!\right)}

\title{Clickbait Detection in YouTube Videos}

\author{Ruchira Gothankar\footnotemark[1]\ \ \ 
Fabio Di Troia\footnotemark[1]\ \ \ 
Mark Stamp\footnotemark[1]\,\,\footnotemark[2]}

\begin{document}

\symbolfootnotetext[1]{Department of Computer Science, San Jose State University}
\symbolfootnotetext[2]{mark.stamp$@$sjsu.edu}

\maketitle

\abstract
YouTube videos often include 
captivating descriptions and intriguing thumbnails designed to increase the number 
of views, and thereby increase the revenue for the person who posted the video. 
This creates an incentive for people to post clickbait videos, in which the content 
might deviate significantly from the title, description, or thumbnail. In effect, 
users are tricked into clicking on clickbait videos. 
In this research, we consider the challenging problem of detecting clickbait
YouTube videos. We experiment with multiple state-of-the-art machine learning 
techniques using a variety of textual features.

\section{Introduction}\label{chap:one}

Today, web content is increasingly popular and people rely on information obtained from the internet. Furthermore, with the diversity of available resources, 
the amount of time spent on the internet has increased. Many platforms 
provide a medium where virtually anyone can publish information that is accessible to a large number of people. However, the credibility of such information is not guaranteed.

Online sources of information include blogs, video sharing platforms, and social media, among others. Many of these applications have been developed with the main intent to generate revenue. Hence, unscrupulous people can use false information to increase their viewership and increase their revenue.
Clickbait is false and deceptive information that lures users to click a link, watch a video, or read an article. It aims to exploit the user's curiosity by providing misleading---though captivating---information. Clickbait has become a marketing tool in many sectors to entice users and thereby to generate revenue. Publishing eye-catching information to manipulate and trick users is a common practice to increase the viewership and spread brand awareness. A clickbait can be an image, a sensational headline, or a misleading video or audio content. While clickbait sources help in gaining attention, there are many disadvantages and negative ramifications. In fact, clickbait not only wastes the time of viewers, but also affects the trustworthiness of the underlying platform~\cite{shang2019towards}.

YouTube is a video publishing platforms where users upload videos and share them with others. When uploading a video, the user adds a title, a description, and a thumbnail. The other users then view the title and thumbnail before deciding whether to view the video. Hence, this data become crucial parameters on which the users can base their decision to watch a video or not. For this reason, many YouTube content creators (aka YouTubers) use clickbait title and thumbnails that might deviate from the actual content to increase viewership for a video, and thereby generate more revenue. 

\begin{figure}[!htb]\centering
\includegraphics[width = 0.7\linewidth]{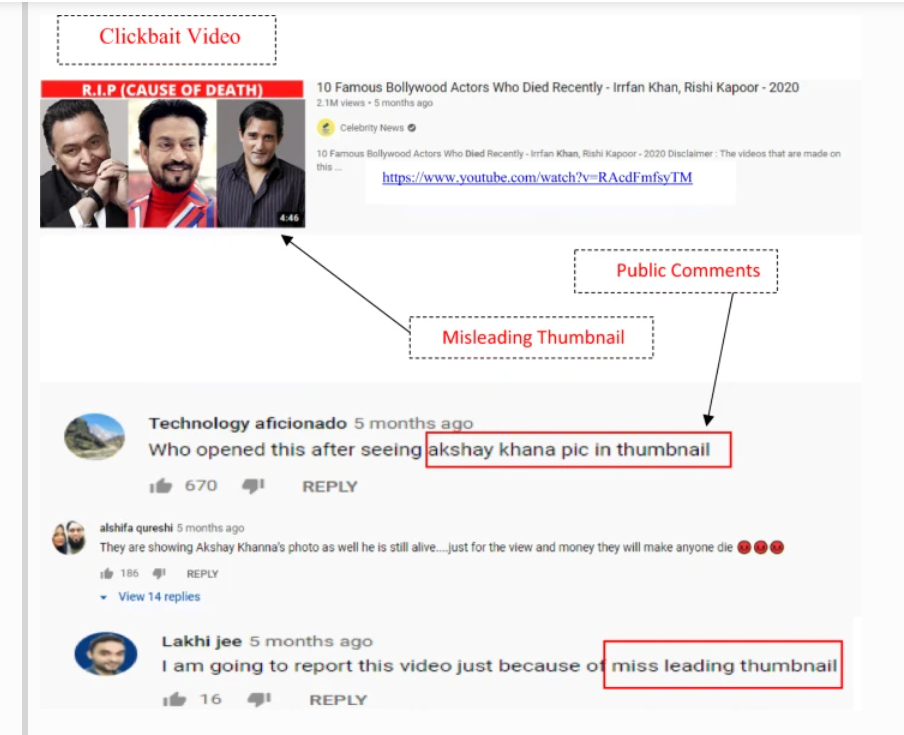}
\caption{Clickbait video example~\cite{varshneyunified}}
\label{fig:clickbait}
\end{figure}

A recent example includes the COVID-19 pandemic, where individuals have posted misleading health-related content, including some fake cures for COVID-19. Some other common examples of clickbait are video titles such as ``You’ll Never Believe What Happened Next$\ldots$'', ``The~10 documentaries you should watch before you die'', ``You Can Now Travel Abroad Without Having to$\ldots$'',``You Won't Believe$\ldots$'' and so on~\cite{hennessey_2020}.
Figure~\ref{fig:clickbait} shows an example of clickbait video on YouTube.

The clickbait problem is somewhat similar to that of spam detection. 
Spam, which is unsolicited emails, often includes misleading messages that are sent to deceive users by redirecting them to websites for the purpose of advertising or attack. Therefore, considerable research has been focused on detecting spam. In this research, we are concerned with detecting clickbait YouTube videos. The YouTube platform relies on users to manually flag 
suspected malicious or clickbait content. However, a more automated approach would 
clearly be desirable. We consider machine learning and deep learning based solutions to 
the clickbait detection problem.

The remainder of this paper is organized as follows. Section~\ref{chap:two} 
considers relevant previous work and background topics related to 
natural language processing (NLP). In Section~\ref{chap:four}, we
discuss our experimental setup, including the datasets used.
Section~\ref{chap:five} contains our experimental results and our 
analysis of these results. In Section~\ref{chap:six}, we give our 
conclusions and we discuss possible directions for future work.

\section{Background}\label{chap:two}

This Section discusses relevant work done in this field. We mainly focuses on clickbait detection, fake news detection, image forgery detection, and hoax detection. Apart from these topics, we also discuss advancements in natural language processing (NLP).

\subsection{Related Work}

Clickbait is a way to attract the attention of the users by luring them to access specific contents. However, misleading information is present on the internet in multiple forms and is often used interchangeably in different contexts. For example, a hoax is spreading false stories of, say, a celebrity death~\cite{varshneyunified}, while an example of a forgery is an image that suggests false information. We now discuss and analyze the performance of previous works on clickbait, fake news, forgery, and hoax detection.

\subsubsection{Clickbait  Detection}

In 2016~\cite{chakraborty2016stop}, Chakraborty et al.~implemented an ML classifier to detect clickbait. They also created a browser extension to help readers navigate around clickbait. They used the headlines from the Wiki-news corpus and used~18,513 articles as legitimate posts. For the clickbait posts, they used articles from popular domains containing illegitimate content. To train their classifier, they used a set of~14 features spanning linguistic analysis, word patterns, and $N$-gram.
They achieved an accuracy of about~89\%\ using a support vector machine (SVM) classifier.

Elyashar et al.~\cite{elyashar2017detecting} developed an approach focused on feature engineering. Their work focused on detecting clickbait posts in online social media. They performed linguistic analysis using a machine learning classifier which could differentiate between legitimate and illegitimate posts. The dataset used for analysis was provided by the 2017 Clickbait Challenge~\cite{click17}. The results of their experiments suggest that malicious content tends to be longer than the benign content. They also concluded that the title of the post played an important role to identify a clickbait.

Glenski et al.~\cite{glenski2017fishing} developed a network model which is a linguistically infused network to detect fake tweets. This model, which is based on long short term memory (LSTM) and convolutional neural networks (CNN), used the text of tweets, images, and description for training. Furthermore, the pretrained embedding model GloVe was used as the embedding 
layer. They achieved an accuracy of~82\%. Zhou~\cite{zhou2017clickbait} proposed a self-attentive neural network model using gated recurrent units (GRU) for predicting fake tweets. They performed multi classification using the annotation scheme. As proof of the success of their approach, they ranked first in the Clickbait Challenge~2017 with an F-score of~0.683.

\subsubsection{Fake News Detection}

Fake news is a type of misinformation that has received considerable attention in recent years. 
The main idea is to analyze the text content of a news item to check if the statements are valid 
or not. Ahmad et al.~\cite{ahmad2020fake} implemented an ensemble model based on the 
linguistic features of the text which involved a combination of multiple machine learning 
algorithms, namely, random forest, multilayer perceptron, and support vector machine (SVM), 
to detect fake news. They used XGBoost as an ensemble learner, achieving an accuracy of~92\%.   

Thota et al.~\cite{thota2018fake} presented a paper on detecting fake news using natural 
language processing. They used TF-IDF and Word2Vec with a dense neural network based 
on the news headline. In another paper on fake news detection, 
Jwa et al.~\cite{jwa2019exbake} implemented a model using bidirectional encoder 
representations from transformers (BERT). The deep contextualizing nature of BERT 
has yielded strong results, including the ability to determine the relationship between 
the  headline and the body of a news article.

\subsubsection{Forgery Detection}

As the name suggests, image forgery detection consists of trying to detect malicious information that is conveyed through images. In 2018, Zhang et al.~\cite{zhang2018fauxbuster} developed a ``fauxtography'' detector which could detect images which are misleading on social media platforms. 

Palod et al.~\cite{palod2019misleading} passed pretrained Word2Vec comment embeddings through an LSTM network to generate a ``fakeness'' vector, and achieved an F-score of~0.82. Shang et al.~\cite{shang2019towards} proposed a model that involved network feature extraction, metadata feature extraction, and linguistic feature extraction to detect clickbait in YouTube videos. The network feature extraction used comments in the videos and extracted semantic features. In the linguistic feature extraction, they relied on document embedding for comments using Doc2Vec, and they also employed a metadata module. In 2019, Reddy et al.~\cite{reddy2019efficient} implemented a model using word embedding and trained on a support vector machine (SVM). In~\cite{dong2019similarity}, 
Dong et al.~have proposed a ``deep similarity-aware attentive model'' that focuses on the relation between the titles that are misleading and the target content. This method was quite different from traditional feature engineering and seemed to work reasonably well. In~\cite{setlur2018semi}, Setlur considered a semi-supervised confidence network along with a gated attention based network. Based on a small labeled dataset, this method gave promising results. 

In many of the above approaches, only the textual information given by the title and the description, along with the metadata features, have been taken into consideration while training a model. An exception is the work in~\cite{shang2019towards}, where the authors have also used comments to extract features. It is also worth noting that the embedding layers of Word2Vec, BERT, and Doc2Vec have been used in all of the implementations mentioned above. 

In this research, we experiment with multiple embedding layers, including BERT, DistilBERT, and Word2Vec. In previous research, BERT has proven to be effective because of its deep contextualizing nature~\cite{jwa2019exbake}. A combination of multiple models, known as ensemble learning, has given interesting results in~\cite{thota2018fake}, and we also consider ensemble models in the form of random forest classifiers. 

\subsubsection{Hoax Detection}

Articles in which facts are knowingly misrepresented can be viewed as hoaxes. These reports provide deceptive information to readers and present it as legitimate facts. One of such examples can be a fake story about a celebrity death. In~\cite{tacchini2017some}, the authors have proposed a technique that uses logistic regression for classifying hoaxes. In the model proposed, they have used features based on user interaction and have achieved an accuracy of 99\%.  Zaman et al.~\cite{zaman2020indonesian} employed a n\"{i}ve Bayes algorithm which uses the feedback from users as an input to verify if a news is a hoax. Kumar et al.~\cite{kumar2016disinformation} have proposed a method which uses random forest classifier to classify the credibility of the articles on Wikipedia. They achieved an accuracy of~92\%. Hoax detection is, though, a less explored area, as compared to the topics discussed above.

\subsection{Natural Language Processing}

\begin{figure}[!htb]\centering
\includegraphics[width = 0.7\linewidth]{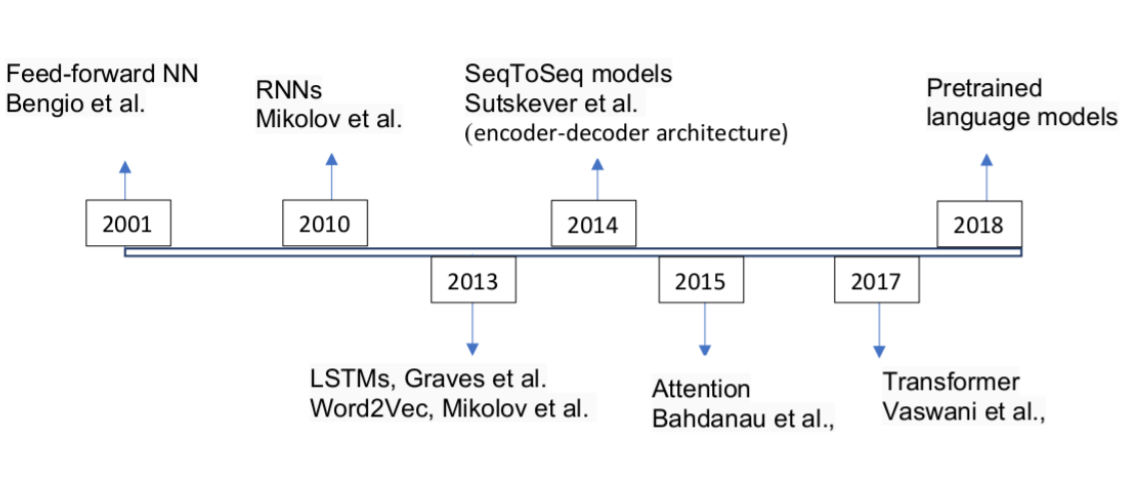}
\caption{NLP advancements in recent years~\cite{rajapaksha2020clickbait}}
\label{fig:advancements}
\end{figure}

Natural language processing (NLP) is the ability of a machine to process and understand the language of a human. It is used to solve many real-world problems, such as machine translation, question answering, and predicting words.
Figure~\ref{fig:advancements} shows a timeline of some recent advances in NLP.

In the early~1990's, statistical and probabilistic approaches were employed to train NLP algorithms. However, with the arrival of the Web, the amount of data grew considerably, and such algorithms became inadequate. In~2001, Bengio et al.~experimented with feedforward neural networks. Later, recurrent neural networks (RNN) and long short-term memory (LSTM) models were introduced~\cite{graves2013generating}. As of~2012, techniques such as latent semantic indexing (LSI), latent semantic analysis (LSA), and support vector machines (SVM) became popular in the NLP domain. Part of speech (POS) tagging is a commonly used approach. 

In 2013, Tomas et. al. introduced Word2Vec, which is used to generate vector
representations of words. These embeddings are obtained from the weights of
a relatively simple neural network, and the vectors can capture important semantic 
information, based on the cosine distance between Word2Vec embeddings~\cite{saifee_2019}. 

Global vector for word representation (GloVe) was introduced in~2014 and is an attempt to combine the benefits of LSA, LSI, and Word2Vec. It is based on the occurrence of a word in the entire corpus. CNNs and LSTMs have become popular for NLP related tasks in recent years, as such models can capture effectively utilize sequential information~\cite{graves2013generating}. LSTM is a highly specialized type of RNN that mitigates the gradient issues that occur with plain vanilla RNNs. Gated recurrent unit (GRU) is a variant of LSTM introduced in~2014 that is lighter, in the sense of having fewer parameters that need to be trained. 

Sutskever et al.~\cite{sutskever2014sequence} proposed a sequence-to-sequence learning approach which uses an encoder-decoder architecture. In fact, such encoder-decoder models appear to be the main language modeling frameworks for NLP tasks today. The concept of an attention mechanism was proposed by Bahdanau et al.~\cite{bahdanau2014neural} in~2015 to overcome the limitation of fixed vector length for input sentences in sequence-to-sequence models~\cite{vaswani2017attention}. Attention provides information about the importance of a part of a sentence during the decision process.

\begin{figure}[!htb]\centering
\includegraphics[width = 0.7\linewidth, height=10cm]{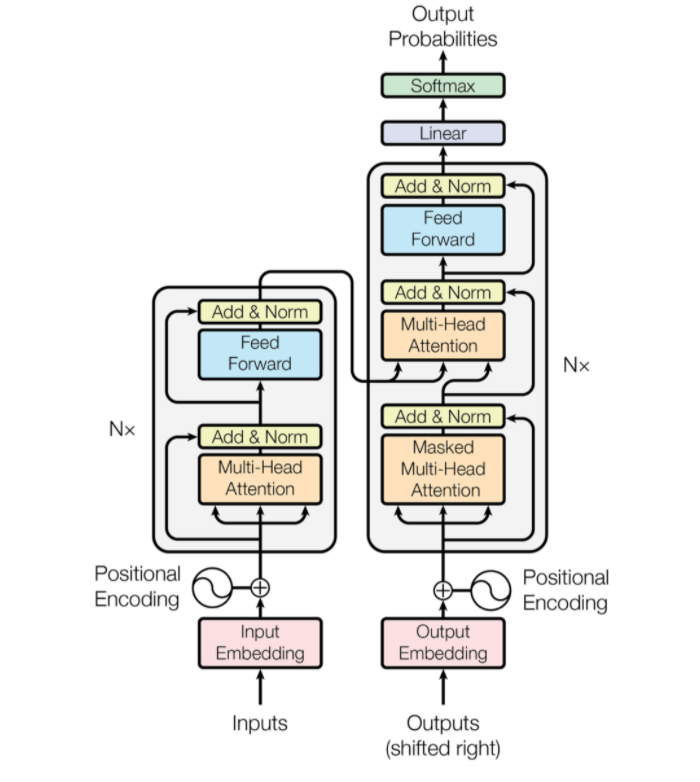}
\caption{Architecture of a transformer~\cite{maxime_2020}}
\label{fig:transformer}
\end{figure}

To better deal with the inherent complexity of attention mechanisms, transformers were introduced~\cite{maxime_2020}. Transformer includes multiple stacks of encoder-decoder architecture, where at each step in the processing, the model takes the output of the previous step as an input. Figure~\ref{fig:transformer} shows the architecture of a transformer where the decoder is on the right and the encoder is on the left. Initially, the input tokens are converted to embedding vectors. Since this model does not have any RNN units, position indices are stored in a $n$-dimensional vector space in the form of embeddings. There are three fully connected layers in this particular attention mechanism, namely, the input key~$K$, the value~$V$, and the query~$Q$, which is a matrix of queries. The algorithm defines weights for words based on all the words in~$K$, and it generates a vector representation for all words based on multi-head attention~\cite{maxime_2020}. The other processes include context fragmentation, and multiple parallel attention layers. Some example of deep learning models that make use of transformers include BERT, RoBERTa, mBERT, and DistilBERT.

\begin{figure}[!htb]\centering
\includegraphics[width = 0.7\linewidth]{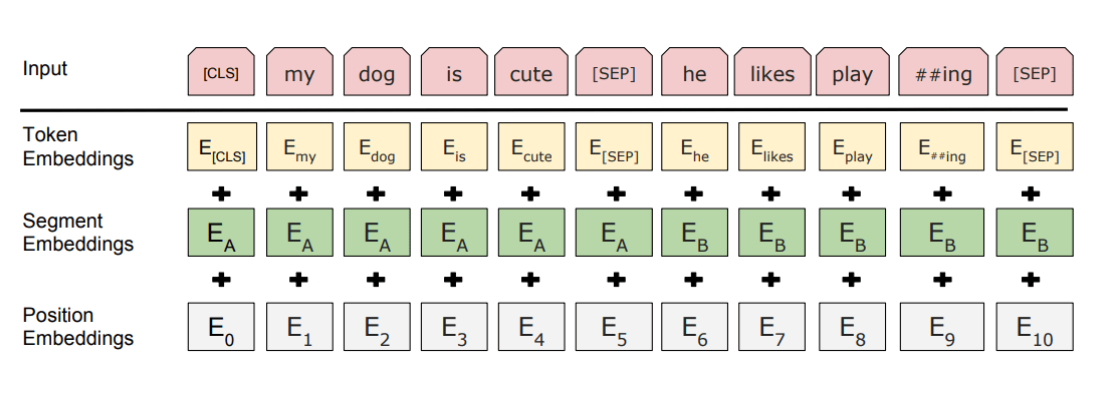}
\caption{Input for the BERT model~\cite{khalid_2019}}
\label{fig:BERT_Input}
\end{figure}

Bidirectional encoder representations from transformer (BERT) uses a transformer which is based on attention to learn the contextual relation between words. It involves an encoder which reads the input, and decoder which predicts the output. It is called bidirectional because instead of reading input sequentially from a specific direction, the transformer reads the sequence of words in both directions. This helps in learning the context of words based on previous and subsequent words.
Figure~\ref{fig:BERT_Input} illustrated the input pattern used in a BERT model.

BERT has four pretrained versions with different layers, hidden nodes, and parameters. Each of these BERT models can be fine-tuned for a specific task by adding additional layers. DistilBERT is a lighter and a faster variant of BERT.

\subsection{Learning Techniques}

In this section, we discuss the various machine learning techniques 
that we have employed in this research. Specifically, we have performed experiments for YouTube clickbait detection based on logistic regression, random forest, and MLP, with various embedding mechanisms.

\subsubsection{Logistic Regression}

Logistic regression is a supervised learning algorithm that is used for categorical data where some parameter---which depend upon the input features and the output---is a categorical prediction. In Logistic regression, a sigmoid function is fitted on the data. 
The formula for the sigmoid function is
\begin{equation}  
\sigma( w^T x + b) = \frac{1}{1 + e^{-(w^T x + b)}}
\label{eq:1:Sigmoid Function}
\end{equation} 
which produces a value in the range of~0 to~1, and hence it can be
interpreted as a probability.
The clickbait detection problem can be treated as a type of binomial logistic regression, 
where the output can be either zero or one~\cite{agrawal_2017}.  

\subsubsection{Random Forest}

A random forest is based on simple decision trees---a large group of decision trees operate together in an ensemble-like manner. Each tree is trained on a subset of the data and features, a process known as boostrap aggregation, or bagging. In bagging, the data for each tree is randomly selected with replacement~\cite{yiu_2019}. 
The final prediction of the random forest can be obtained via a simple voting scheme. 
A random forest mitigates the tendency of individual decision trees to overfit the training data. 

The important hyperparameters in a random forest are $n$-estimators, $n$-jobs, max-features, and min-sample-leaf. The $n$-estimators parameter represent the number of trees that are constructed. Typically, adding more trees increases performance at the cost of computation time. The max-features parameter is the number of features required to split at a specific node. The 
parameter $n$-jobs is the number of processors that work in parallel. 

\subsubsection{Multilayer Perceptron}

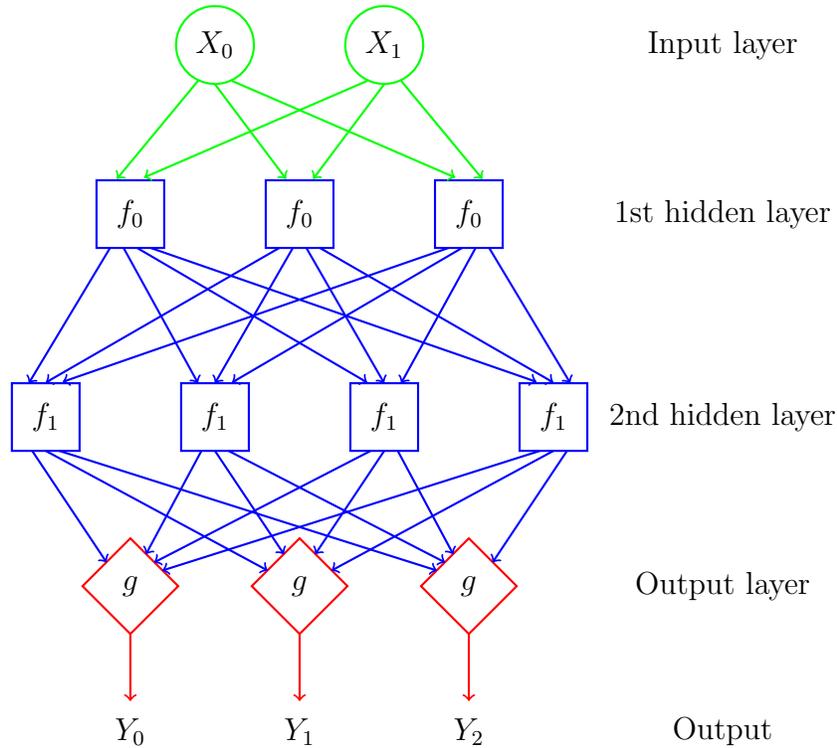
\begin{figure}[!htb]
\centering
\begin{tikzpicture}[scale=0.9]
    
    % circles
    \draw[thick,color=green] (3.0,8.5) circle (0.575);
    \draw[thick,color=green] (5.5,8.5) circle (0.575);

    % squares (top)
    \draw[thick,color=blue] (1.25,5.5) rectangle (2.25,6.5);
    \draw[thick,color=blue] (3.75,5.5) rectangle (4.75,6.5);
    \draw[thick,color=blue] (6.25,5.5) rectangle (7.25,6.5);
    
    % squares (bottom)
    \draw[thick,color=blue] (0.0,2.5) rectangle (1.0,3.5);
    \draw[thick,color=blue] (2.5,2.5) rectangle (3.5,3.5);
    \draw[thick,color=blue] (5.0,2.5) rectangle (6.0,3.5);
    \draw[thick,color=blue] (7.5,2.5) rectangle (8.5,3.5);
    
    % diamonds
    \draw[thick,color=red,rotate around={45:(1.75,0.5)}] (1.25,0.0) rectangle (2.25,1.0);
    \draw[thick,color=red,rotate around={45:(4.25,0.5)}] (3.75,0.0) rectangle (4.75,1.0);
    \draw[thick,color=red,rotate around={45:(6.75,0.5)}] (6.25,0.0) rectangle (7.25,1.0);

    % circle to square
    \draw[thick,color=green,->] (2.75,7.97) -- (1.55,6.52);
    \draw[thick,color=green,->] (3,7.92) -- (4.05,6.52);
    \draw[thick,color=green,->] (3.25,7.97) -- (6.55,6.54);
    
    \draw[thick,color=green,->] (5.25,7.97) -- (1.95,6.54);
    \draw[thick,color=green,->] (5.5,7.92) -- (4.45,6.52);
    \draw[thick,color=green,->] (5.75,7.97) -- (6.95,6.52);

    % square to diamond
    \draw[thick,color=blue,->] (0.3,2.5) -- (1.4,0.85);
    \draw[thick,color=blue,->] (0.5,2.5) -- (3.78,0.73);
    \draw[thick,color=blue,->] (0.7,2.5) -- (6.28,0.73);
    
    \draw[thick,color=blue,->] (2.8,2.5) -- (1.98,0.97);
    \draw[thick,color=blue,->] (3.0,2.5) -- (4.02,0.97);
    \draw[thick,color=blue,->] (3.2,2.5) -- (6.4,0.85);
    
    \draw[thick,color=blue,->] (5.3,2.5) -- (2.1,0.85);
    \draw[thick,color=blue,->] (5.5,2.5) -- (4.48,0.97);
    \draw[thick,color=blue,->] (5.7,2.5) -- (6.52,0.97);
    
    \draw[thick,color=blue,->] (7.8,2.5) -- (2.22,0.73);
    \draw[thick,color=blue,->] (8.0,2.5) -- (4.72,0.73);
    \draw[thick,color=blue,->] (8.2,2.5) -- (7.1,0.85);

    % square to square
    \draw[thick,color=blue,->] (1.45,5.5) -- (0.25,3.5);
    \draw[thick,color=blue,->] (1.65,5.5) -- (2.75,3.5);
    \draw[thick,color=blue,->] (1.85,5.5) -- (5.25,3.51);
    \draw[thick,color=blue,->] (2.05,5.5) -- (7.75,3.52);

    \draw[thick,color=blue,->] (3.95,5.5) -- (0.5,3.51);
    \draw[thick,color=blue,->] (4.15,5.5) -- (3.0,3.5);
    \draw[thick,color=blue,->] (4.35,5.5) -- (5.5,3.5);
    \draw[thick,color=blue,->] (4.55,5.5) -- (8.0,3.51);

    \draw[thick,color=blue,->] (6.45,5.5) -- (0.75,3.52);
    \draw[thick,color=blue,->] (6.65,5.5) -- (3.25,3.51);
    \draw[thick,color=blue,->] (6.85,5.5) -- (5.75,3.5);
    \draw[thick,color=blue,->] (7.05,5.5) -- (8.25,3.5);

    % output
    \draw[thick,color=red,->] (1.75,-0.2) -- (1.75,-1.2);
    \draw[thick,color=red,->] (4.25,-0.2) -- (4.25,-1.2);
    \draw[thick,color=red,->] (6.75,-0.2) -- (6.75,-1.2);

    % labels for circles
    \node at (3.0,8.5) {$X_0$};
    \node at (5.5,8.5) {$X_1$};

    % labels for squares (top)
    \node at (1.75,6.0) {$f_0$};
    \node at (4.25,6.0) {$f_0$};
    \node at (6.75,6.0) {$f_0$};

    % labels for squares (bottom)
    \node at (0.5,3.0) {$f_1$};
    \node at (3.0,3.0) {$f_1$};
    \node at (5.5,3.0) {$f_1$};
    \node at (8.0,3.0) {$f_1$};

    % labels for diamonds
    \node at (1.75,0.5) {$g$};
    \node at (4.25,0.5) {$g$};
    \node at (6.75,0.5) {$g$};

    % labels for output
    \node at (1.75,-1.65) {$Y_0$};
    \node at (4.25,-1.65) {$Y_1$};
    \node at (6.75,-1.65) {$Y_2$};
    
    % labels
    \node at (10.5,8.5) {Input layer};
    \node at (10.5,6.0) {1st hidden layer};
    \node at (10.5,3.0) {2nd hidden layer};
    \node at (10.5,0.5) {Output layer};
    \node at (10.5,-1.65) {Output};

\end{tikzpicture}
\caption{MLP with two hidden layers}\label{fig:multiLayerPerceptron}
\end{figure}

A multilayer perceptron (MLP) is a basic type of feedforward neural network
that includes input and output layers, along with at least one hidden layer. 
An MLP with two hidden layers is illustrated in Figure~\ref{fig:multiLayerPerceptron}.

The output layer of an MLP can be used for prediction or classification. 
Next, we briefly discuss 
regularization and activation functions; see~\cite{franckepeixoto_2020}
for additional details on these and related topics.

Neural network models are prone to overfitting. An overfitted model is very effective in classifying the training data but it obtains poor accuracy in predicting the test data---in effect, the model has
``memorized'' the training data, rather than learning from the training data.
One useful technique to prevent overfitting is the use of dropouts, 
where some number of nodes are ignored during various training 
steps~\cite{franckepeixoto_2020}. This simple approach forces nodes that
would otherwise atrophy to become active in the learning process.

An activation function is used to determine the output of node in a neural network. 
There are multiple types of activation functions, including tanh, sigmoid, ReLU, 
and leaky ReLU~\cite{franckepeixoto_2020}. In this research, we have experimented 
with ReLU and tanh. 

\section{Implementation}\label{chap:four}

This section includes details on the implementation used in this research. 
We discuss the setup used to train and execute the various machine learning models, 
the experimental design, and so on.

\subsection{Hardware and Software}

In this research, we used multiple Conda virtual environments for each implementation. Conda is an open source package and environment management system which runs on multiple operating systems~\cite{conda}. The host machine was configured as given in Table~\ref{tab:config}.
All the training and the experiments were run on the host machine. 

\begin{table}[!htb]
\centering
\caption{Host machine configuration}\label{tab:config}
\adjustbox{scale=0.85}{
\begin{tabular}{ll}
\midrule\midrule
Component & Details \\ \midrule
Model & ASUS ZenBook \\
Processor & Intel(R) Core(TM) i7-8565U CPU @ 1.80GHz   1.99 GHz \\
RAM & 16.0 GB \\
System Type & 64-bit OS \\
Operating System & Windows 10 \\
\midrule\midrule
\end{tabular}
}
\end{table}

\subsection{Approach}

Our clickbait detection experiments are based on a set of labeled videos. 
The problem is formulated as a binary classification problem where for each video a
machine learning algorithm classifies it is clickbait or non-clickbait.
The information from multiple sources (e.g., title, description, comments) 
are combined and fed to the classification model. The performance is evaluated and analyzed by multiple measures, specifically, precision, recall and the F-score. 

\begin{figure}[!htb]\centering
\includegraphics[width = 0.7\linewidth]{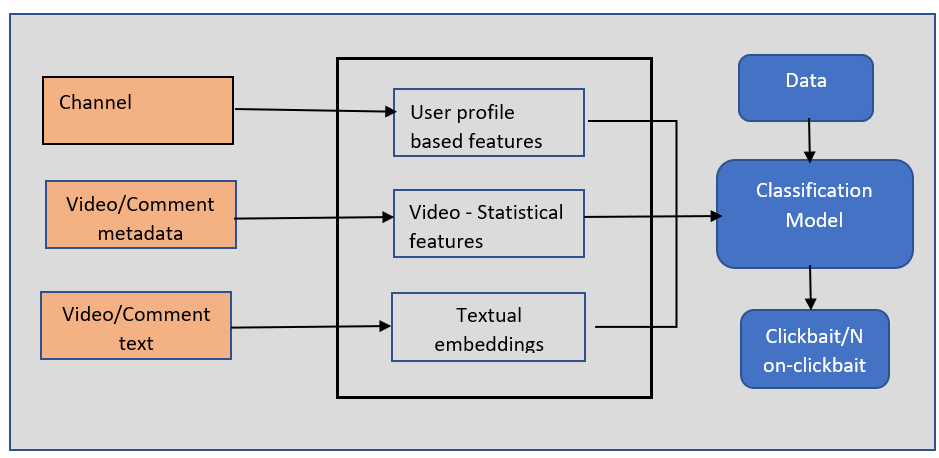}
%%%%% Need to be consistent wrt capitalization of captions
\caption{Overview of clickbait classification model}
\label{fig:fig1}
\end{figure}

There are three types of features considered in this research. The first involves features from the profile of the user who posted the video (subscriptions count, views count, and videos count). 
The second type of feature is based on extracting textual information from the video (title and description). The third component involves statistical features related to the video 
(like count, dislike count, like-dislike ratio, views, and number of comments).
A classification model performs binary classification (clickbait or non-clickbait) 
based on some combination of these features. An overview is provided in Figure~\ref{fig:fig1}.

\subsection{Features}

Features that provide information regarding the reputation of the channel and the videos include the number of subscribers of the YouTube channel, the number of likes or upvotes, and the age of the channel. These statistical features represent the response of viewers to the channel. Previous related work claimes that videos that are clickbait tend to have a relatively small number of subscribers and likes~\cite{varshneyunified}.

Usually, the number of views for the clickbait and non-clickbait videos are quite similar~\cite{zannettou2018good}. Useful information in determining the credibility of a video is given by the dislike ratio, the favorites count, the video age, the views count, and the comments count. Sometime, in clickbait videos the uploader disables the comment section. This itself provides clues about the video~\cite{varshneyunified}.
 
Textual features include the headline of the video, the description of the video, and the comments by the viewers.
YouTubers who upload clickbait usually employ techniques which are deceptive. 
They use catchy and exaggerated phrases for the title and description of the video. Some common phrases are ``viral,'' ``top.'' ``won't believe'', ``epic'', and similar. We tokenize the text and embed 
it in classification models using various embedding techniques, including Word2Vec, BERT, and DistilBERT 

\subsection{Dataset}

Every month, billions of people visit YouTube and the videos are watched for over a billion hours. A large number of videos are also uploaded by the users. In fact, YouTube is a platform where people can generate revenue by uploading videos and gaining viewership for their videos. 

In this research, the evaluation is done on a dataset of~8219 labeled videos, 
where~4300 are non-clickbait and~3919 are clickbait. The dataset was crawled from the
Google YouTube API for the list of video IDs fetched from a Github source~\cite{alessiovierti}. These sources were randomly picked and manually verified. 
The statistics for various parameters are shown in Table~\ref{tab:1}.

\begin{table}[!htb]
\caption{Dataset statistics}\label{tab:1}
\centering
\adjustbox{scale=0.85}{
\begin{tabular}{l|rrr}\midrule\midrule 
Data Item          & Min   & Mean      & Max          \\ \midrule
Title length       & 10    & 54        & 107          \\
Description length & 15    & 1131      & 5162         \\
View count         & 21    & 5,660,978   & 2,543,466,463   \\
Comment count      & 0     & 522       & 49,060        \\
Like count         & 0     & 49,615     & 13,542,232     \\
Subscriber count   & 977   & 10,200     & 23,695,417     \\
Dislike count      & 0     & 1320      & 516,171    \\ \midrule\midrule
\end{tabular}
}
\end{table}

\subsection{Experiments}

In this research, we experimented multiple techniques including multiple language modeling techniques. We used Word2Vec, BERT, and DistilBERT for word embeddings. Architecture for the individual models is also shown. A grid search was used for training and building the models to obtain the best set of parameters. In this section, we briefly describe each of our models, and in the next section
we give the results for each experiment.

\subsubsection{Experiment~I: Logistic Regression with Word2Vec}

In this experiment, we used a Word2Vec model provided by Gensim~\cite{gensim} to 
generate the vector representations of words in the dataset. A logistic regression model is trained on these embeddings along with additional features, specifically, comments count, likes count, dislikes count, and subscriptions count for the channel.

\subsubsection{Experiment~II: Random Forest with Word2Vec}

In this experiment, a random forest classifier is trained on the Word2Vec embeddings. We again used the Word2Vec model provided by Gensim. The values tested for $n$-estimators is!10, 20, 30, 50, 
and~100. The set of input features are title, description, and metadata features such as comments count, likes count, dislikes count, and subscriptions count for the channel.

\subsubsection{Experiment~III: MLP with Word2Vec}

In this case, we again use the Word2Vec model provided by Gensim. T
The embedding for title and description is concatenated with the metadata features of the video and is fed to an MLP for classification. The batch size is~10 for~40 epochs. The activation functions used are ReLU and sigmoid. 
%Figure~\ref{fig:wwchart} shows the overall architecture of the model. 
Figure~\ref{fig:wwchart} in the appendix provides the overall architecture of the model. 

%\begin{figure}[!htb]\centering
%\includegraphics[width = 0.65\linewidth]{images/ww/chart.png}
%\caption{Architecture of model for experiment~III}
%\label{fig:wwchart}
%\end{figure}

Note that we use two input embedding layers for textual data (namely, title and description), which are then concatenated together. After this step, the output from the dense layer is flattened and concatenated with the input for the metadata features. Finally, a fully connected layer is used for classification.

\subsubsection{Experiment~IV: MLP with DropOut, Batch Normalization, and Word2Vec}

This experiment is an optimization of the previous experiment. In this model, 
additional dense layers, along with batch normalization and dropout rate of~0.5, are employed. We have used parametric rectified linear units (PReLU) as the activation function. The batch size is
again~10 for~40 epochs. 
%Figure~\ref{fig:wv1Chart} represents the overall architecture of the model. 
Figure~\ref{fig:wv1Chart} in the appendix illustrates the overall architecture of the model. 

%\begin{figure}[!htb]\centering
%\includegraphics[width = 0.65\linewidth]{images/wv1/chart.png}
%\caption{Architecture of model for experiment~IV}
%\label{fig:wv1Chart}
%\end{figure}

In this model, the output from the embedding layers for the textual data is concatenated, followed by a fully connected dense layer, batch normalization, and activation. This output is finally flattened and concatenated with the metadata features. 

\subsubsection{Experiment~V: MLP with BERT}

In this experiment, we have used BERT embedding for title and description of the video. The advantage of using BERT as an embedding model is that it provides context-based representation for each word in a sentence. In contrast, Word2Vec provides representations which are fixed irrespective of where the word is used in the sentence. The pretrained model of BERT that is used in this experiment has~12 layers, 110M parameters, and~768 hidden layers. The BERT tokenizer is used to split the words into tokens and attention masks are used for padding. The mask value of one is for tokens that are not masked, while the value zero means that the token is added by padding and should not be considered for attention. The model uses Adam optimizer, and the batch size is~10 for~5 epochs, and we have used sequence of length~180 for this experiment. 
%Figure~\ref{fig:bertChart} shows the model architecture. 
Figure~\ref{fig:bertChart} in the appendix shows the model architecture. 

%\begin{figure}[!htb]\centering
%\includegraphics[width = 0.65\linewidth]{images/BERT/chart.png}
%\caption{Architecture of model for experiment~V}
%\label{fig:bertChart}
%\end{figure}

Note that the output of the BERT embedding layer is followed by a dense layer, which is then concatenated with the metadata features. After this, a dropout layer followed by a fully connected layer is used for classifying the data.

\subsubsection{Experiment~VI: MLP with DistilBERT}

DistilBERT is a faster, lighter model that is a variant of BERT---it runs~60\% faster and has~45\% fewer parameters than BERT~\cite{distilbert_documentation}. For this experiment, we have used a pretrained DistilBERT model. The embeddings for tile and description are fed into a MLP and, later, concatenated with the metadata features of the video and the YouTube channel. The model uses Adam optimizer and the batch size is~10 for~5 epochs. 
%Figure~\ref{fig:distilchart} shows the model architecture. 
Figure~\ref{fig:distilchart} in the appendix gives details on this model architecture. 

%\begin{figure}[!htb]\centering
%\includegraphics[width = 0.65\linewidth]{images/Distil/chart.png}
%\caption{Architecture of model for experiment~VI}
%\label{fig:distilchart}
%\end{figure}

Note that in this model, the input from the metadata features is concatenated.
Of course, the output layer is a dense layer that is used for classification.

\section{Results}\label{chap:five}

\begin{figure}[!htb]\centering
\includegraphics[width = 0.55\linewidth]{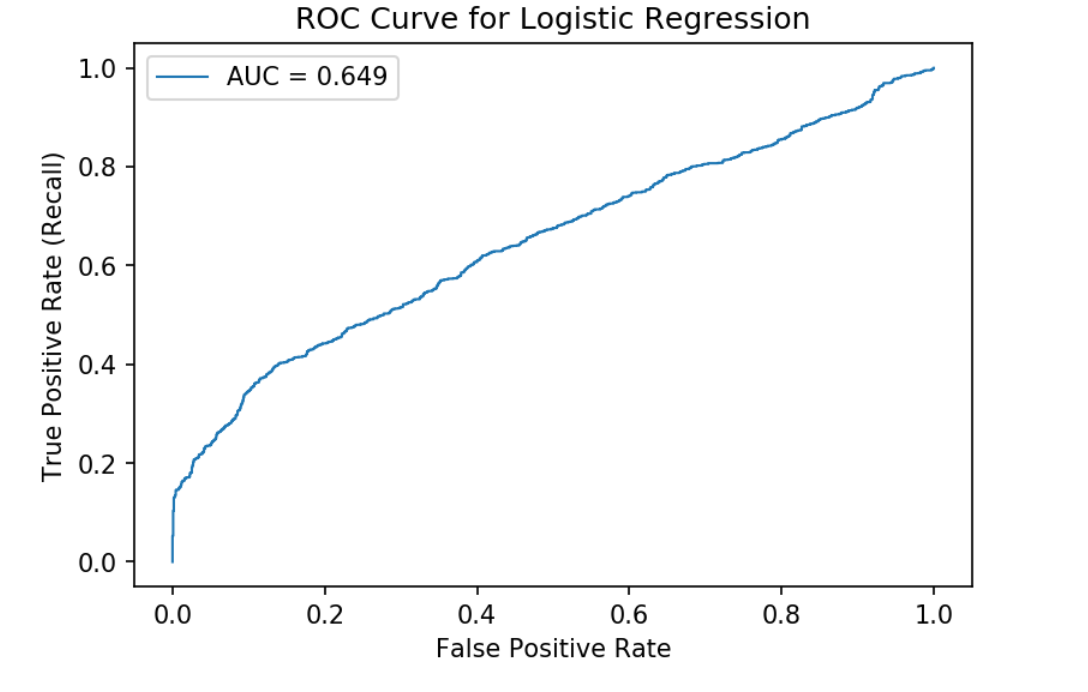}
\caption{ROC curve for logistic regression}
\label{fig:lroc}
\end{figure}

Recall that in experiment~I, we have use logistic regression with Word2Vec embeddings for the features title and description, along with the metadata features. In this case, we achieve an accuracy of 52\% with just title as input, and an accuracy of 70\% with all of these features. This model is fast to train and much simple to implement. Figure~\ref{fig:lroc} shows the ROC curve for this logistic regression model.

Experiment~II involves using a random forest classifier based on the title, description, likes count, dislikes count, comments count, and subscriptions count. We used Word2Vec embeddings for title and description. We trained this model in multiple sets of inputs. The first set of inputs includes just the title and metadata features. The last set of inputs included all the features. Not surprisingly, we find that the accuracy improves as more features are added. 

Table~\ref{table:classification:report1} shows precision and accuracy of 80.1\% for the model with the first set of input features, that is, title and two metadata features for likes count and dislikes count.

%%%%% Use a consistent table format, such as this one
\begin{table}[!htb]
\caption{Random forest based on title and like/dislike counts}
\label{table:classification:report1}
\centering
\adjustbox{scale=0.85}{
\begin{tabular}{c | c c c r}\midrule\midrule
Class & Precision & Recall & F-score & Support\\
\midrule
non-clickbait & 0.81 & 0.80 & 0.81 & 1275\\
clickbait & 0.80 & 0.81 & 0.80 & 1182\\
\midrule
accuracy & --- & --- & 0.80 & 2457\\
macro avg & 0.80 & 0.80 & 0.80 & 2457\\
weighted avg & 0.80 & 0.80 & 0.80 & 2457\\ 
\midrule\midrule
\end{tabular}
}
\end{table}

Table~\ref{table:classification:report2} shows the report for this experiment when we use the title and all the metadata as features. The accuracy for this experiment is~92.5\%. The report shows the precision and recall of the model in classifying clickbait and non-clickbait videos. The model performs slightly better in classifying non-clickbait videos.

\begin{table}[!htb]
\caption{Random forest based on title and metadata features}
\label{table:classification:report2}
\centering
\adjustbox{scale=0.85}{
\begin{tabular}{c | c c c r}\midrule\midrule
Class & Precision & Recall & F-score & Support\\
\midrule
non-clickbait & 0.93 & 0.93 & 0.93 & 1275\\
clickbait & 0.92 & 0.92 & 0.92 & 1182\\
\midrule
accuracy & --- & --- & 0.93 & 2457\\
macro avg & 0.93 & 0.93 & 0.93 & 2457\\
weighted avg & 0.93 & 0.93 & 0.93 & 2457\\
\midrule\midrule
\end{tabular}
}
\end{table}

\begin{figure}[!htb]\centering
\includegraphics[width = 0.55\linewidth]{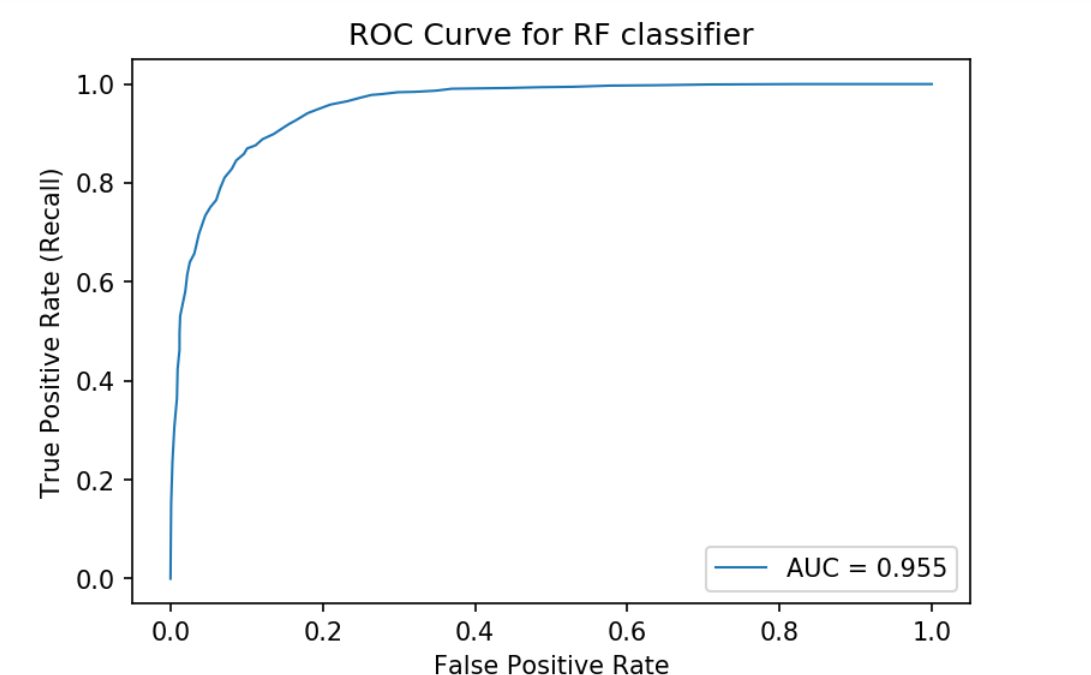}
\caption{ROC curve for random forest}
\label{fig:rocRandom}
\end{figure}

Figure~\ref{fig:rocRandom} shows the ROC curve for the random forest model where the input features included title, description, and all the metadata features, that is, count, dislikes count, comments count, subscriptions count, and views count. The AUC for this model is~0.95 with an accuracy of~94\%. This shows that the model performs well and that adding more features increases the accuracy of the model. 

\begin{figure}[!htb]
\centering
\begin{tabular}{cc}
\includegraphics[width = 0.4\linewidth]{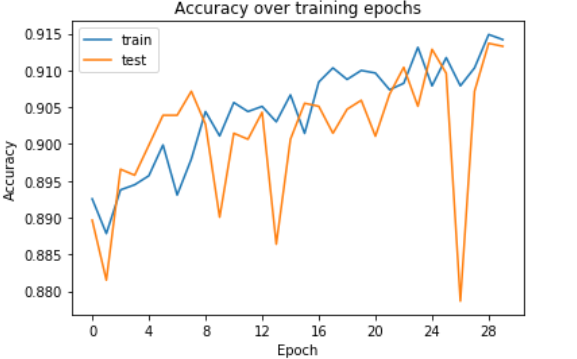}
&
\includegraphics[width = 0.38\linewidth]{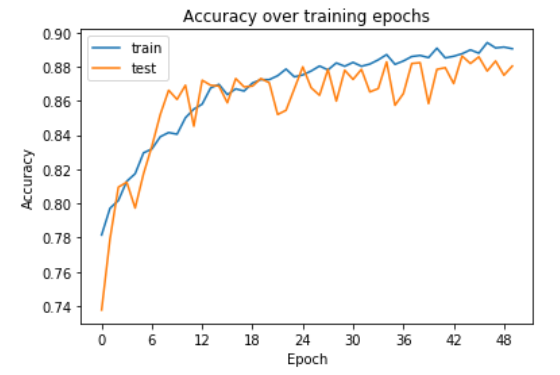}
\\
(a) Experiment~III
&
(b) Experiment~IV
\\
\\
\includegraphics[width = 0.38\linewidth]{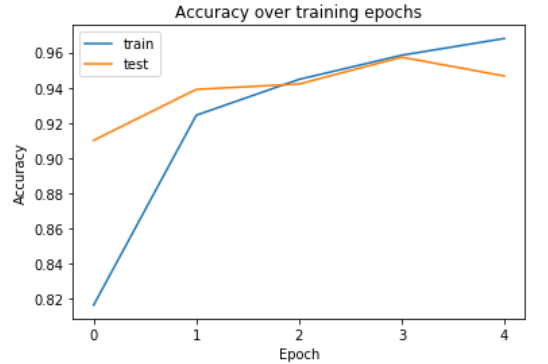}
&
\includegraphics[width = 0.4\linewidth]{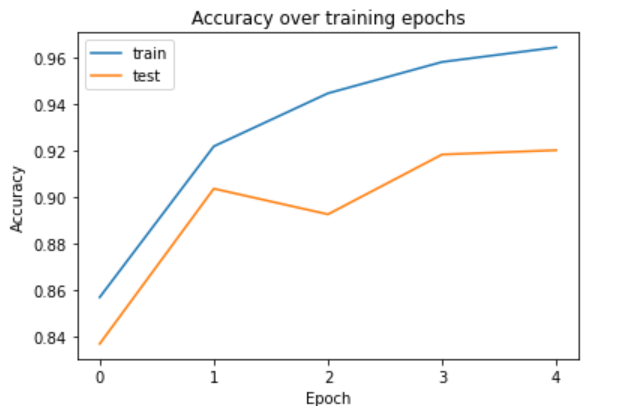}
\\
(c) Experiment~V
&
(d) Experiment~VI
\end{tabular}
\caption{Accuracy over training epochs for experiments~III through~VI}
\label{fig:accuracy36}
\end{figure}

In experiment~III a simple MLP is used for classification, based on Word2Vec embeddings for title and description that are concatenated with metadata features. In this case, the test accuracy is observed to fluctuate during the training process, but the best average accuracy achieved is better than~91\%. Figure~\ref{fig:accuracy36}~(a) shows the accuracy for this experiment over the~30 training epochs. 

%\begin{figure}[!htb]\centering
%\includegraphics[width = 0.65\linewidth]{images/ww/accuracy.png}
%\caption{Accuracy over training epochs for experiment~III (MLP with Word2Vec)}
%\label{fig:wwaccuracy}
%\end{figure}

In experiment~IV, a modified MLP is used with batch normalization and PReLU as an activation function. In this case, the accuracy is slightly worse than in experiment~III, although
the training is more stable, as can be observed in Figure~\ref{fig:accuracy36}~(b).

%\begin{figure}[!htb]\centering
%\includegraphics[width = 0.65\linewidth]{images/wv1/accuracy.png}
%\caption{Accuracy over training epochs for experiment~IV (MLP with Word2Vec,
%BatchNorm, and ReLU)}
%\label{fig:wv1accuracy}
%\end{figure}

In experiment~V, we have used a transfer learning model based on BERT for word embeddings. This experiment with BERT gives an accuracy of~94.5\%. In this experiment the length of the input sequence is fixed at~180 characters. Figure~\ref{fig:accuracy36}~(c) shows the plot for accuracy over training epochs for both the train and validation sets. Note that the number of epochs is small
due to the extended training time required, as compared to other models considered.

%\begin{figure}[!htb]\centering
%\includegraphics[width = 0.65\linewidth]{images/BERT/accuracy.png}
%\caption{Accuracy over training epochs for experiment~V (MLP with BERT)}
%\label{fig:BERTaccuracy}
%\end{figure}

In experiment~VI, we have used a lighter variant of BERT model for the word embeddings, namely, DistilBERT. The accuracy achieved in this case is around~92\%. 
This model is significantly faster to train than the BERT, 
although the accuracy obtained with BERT is slightly better than using DistilBERT. 
Table~\ref{table:distilclassification_report} shows the precision and recall for experiment~VI, 
while Figure~\ref{fig:accuracy36}~(d) shows the training and test accuracy over epochs. 

\begin{table}[!htb]
\caption{Results for experiment~VI (MLP with DistilBERT)}
\label{table:distilclassification_report}
\centering
\adjustbox{scale=0.85}{
\begin{tabular}{c | c c c r}\midrule\midrule
Class & Precision & Recall & F-score & Support\\
\midrule
non-clickbait & 0.92 & 0.95 & 0.93 & 884\\
clickbait & 0.93 & 0.89 & 0.91 & 754\\
\midrule
accuracy & --- & --- & 0.92 & 1638\\
macro avg & 0.92 & 0.92 & 0.92 & 1638\\
weighted avg & 0.92 & 0.92 & 0.92 & 1638\\
\midrule\midrule
\end{tabular}
}
\end{table}
  
%\begin{figure}[!htb]\centering
%\includegraphics[width = 0.65\linewidth]{images/Distil/accuracy.png}
%\caption{Accuracy over training epochs for experiment~VI (MLP with DistilBERT)}
%\label{fig:distilaccuracy}
%\end{figure}

\begin{figure}[!htb]
\centering
\begin{tikzpicture}[scale=0.95, every node/.style={scale=1.0}]
\begin{axis}[%bar shift=0pt,
        width  = 0.6*\textwidth,
        height = 7.5cm,
        ymin=0.0,ymax=1.125,
        ytick={0.0,0.2,0.4,0.6,0.8,1.0},
        major x tick style = transparent,
        ybar=5*\pgflinewidth,
        bar width=18.0pt,
%        ymajorgrids = true,
        ylabel = {Accuracy},
        symbolic x coords={
        		Logistic regression with Word2Vec, 
        		Random forest with Word2Vec, 
        		MLP with Word2Vec, 
		MLP with Word2Vec and dropout, 
		MLP with BERT, 
		MLP with DistilBERT},
	xticklabels={
		{Logistic regression (Word2Vec)}, 
        		{Random forest (Word2Vec)},
        		{MLP (Word2Vec)}, 
		{MLP plus (Word2Vec)}, 
		{MLP (BERT)}, 
		{MLP (DistilBERT)}},
	y tick label style={
    		/pgf/number format/.cd,
   		fixed,
   		fixed zerofill,
    		precision=2},
%	yticklabel pos=right,
        xtick = data,
        x tick label style={
        		rotate=60,
		font=\small,
		anchor=north east,
		inner sep=0mm},
%		font=\small},
%        scaled y ticks = false,
	%%%%% numbers on bars and rotated
        nodes near coords,
        every node near coord/.append style={rotate=90, 
        								   anchor=west,
								   font=\footnotesize,
								   /pgf/number format/.cd,
								   fixed,
								   fixed zerofill,
								   precision=2},
        %%%%%
%        enlarge x limits=0.03,
        enlarge x limits=0.1,
        legend cell align=left,
        legend style={
%                at={(1,1.05)},
%                anchor=south east,
%	        nodes={rotate=90},%%%%% rotate text in legend
%                at={(0.125,0)},
%                at={(0.125,0)},
%                at={(0.8775,0)},
                at={(0.89,0.02)},
                anchor=south,
                column sep=1ex
        },
]
\addplot [fill=blue,opacity=1.00]
coordinates {
(Logistic regression with Word2Vec, 0.7000)
(Random forest with Word2Vec, 0.9250)
(MLP with Word2Vec, 0.9125)
(MLP with Word2Vec and dropout, 0.8800)
(MLP with BERT, 0.9450)
(MLP with DistilBERT, 0.9200)
};
\end{axis}
\end{tikzpicture}
\caption{Accuracy comparison of experiments}
\label{fig:compare}
\end{figure}
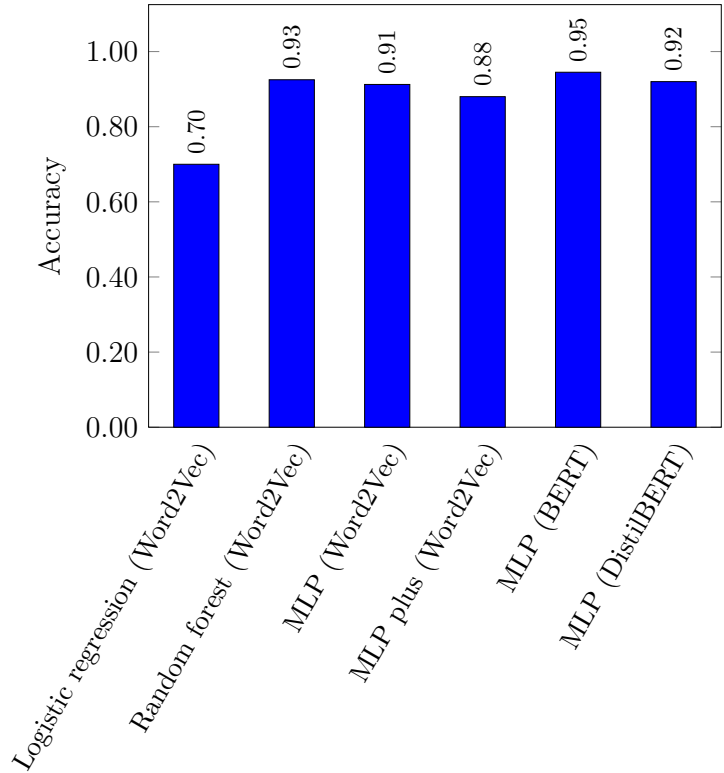

In Figure~\ref{fig:compare} we summarize the results of our
six experiments in terms of accuracy (to two decimal places).
Note that in the bar graph in Figure~\ref{fig:compare},
``MLP plus'' is used to denote our MLP model
that includes dropout and batch normalization.
Also, the bars from left-to-right represent experiments~I through~VI,
respectively.

\section{Conclusion and Future Works}\label{chap:six}  

The goal of this research was to utilize state-of-the-art techniques to classify YouTube videos as clickbait or non-clickbait. A YouTube video has multiple characteristics that can serve as useful features for such classification. We leverage three main types of such features, namely, user profile, video statistics, and textual data. In this research, multiple classification techniques were considered, including logistic regression, random forest, and MLP, and we employed Word2Vec, BERT, and DistilBERT as language models. The best accuracy was achieved using an MLP classifier based on BERT embeddings, but a the more lightweight DistilBERT performed almost as well. We also confirmed that the accuracy of the models could be increased by adding more features.

For future work, more features can be included. For instance, the transcript of the video might contain useful information. For example, the ``distance'' between the transcripts and the title could provide important insight, as the content of clickbait videos often differs significantly from the title. The network structure of the comments and replies, which represents the semantic features and attributes, can also be considered~\cite{shang2019towards}.

In this research, we experimented with BERT, Word2Vec, and DistilBERT for word embeddings. For future work, DocToVec embeddings could also be considered. We used random forest classifier, and other ensemble techniques could be considered, including, such as XGBoost.
Furthermore, we can also experiment with state-of-the-art attentive language models, such as XLNet, which is supposed to be better than BERT for determining long-term dependencies~\cite{tum_2020}.

\bibliographystyle{plain}
\bibliography{references.bib}

\clearpage

\appendix

\section*{Appendix: Model Architectures}

\begin{figure}[!htb]\centering
\includegraphics[width = 0.9\linewidth]{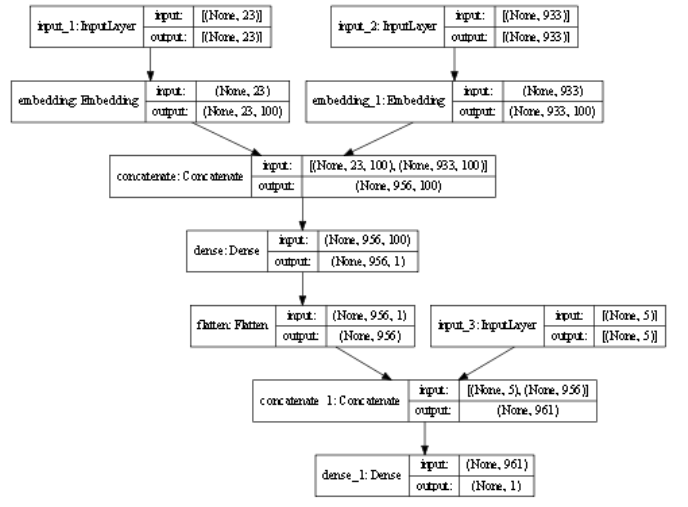}
\caption{Architecture of model for experiment~III}\label{fig:wwchart}
\end{figure}

\begin{figure}[!htb]\centering
\includegraphics[width = 0.9\linewidth]{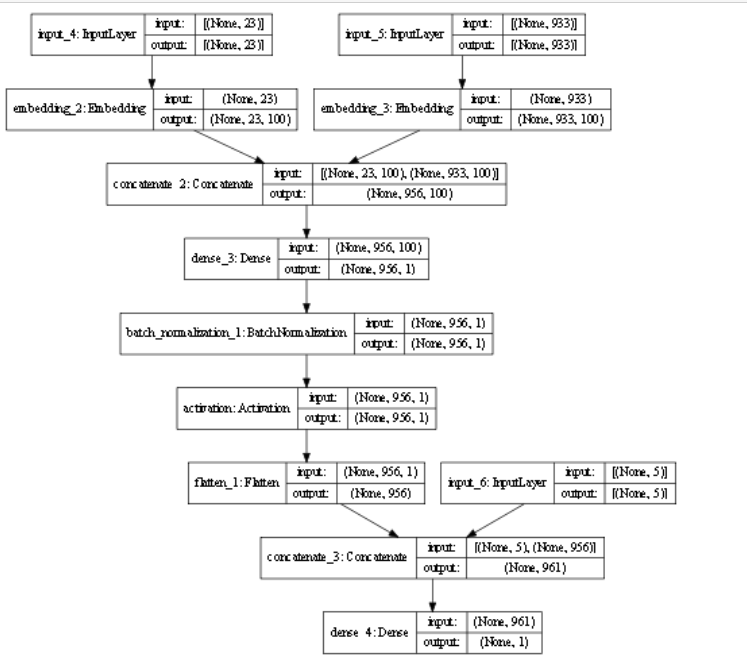}
\caption{Architecture of model for experiment~IV}\label{fig:wv1Chart}
\end{figure}

\begin{figure}[!htb]\centering
\includegraphics[width = 0.9\linewidth]{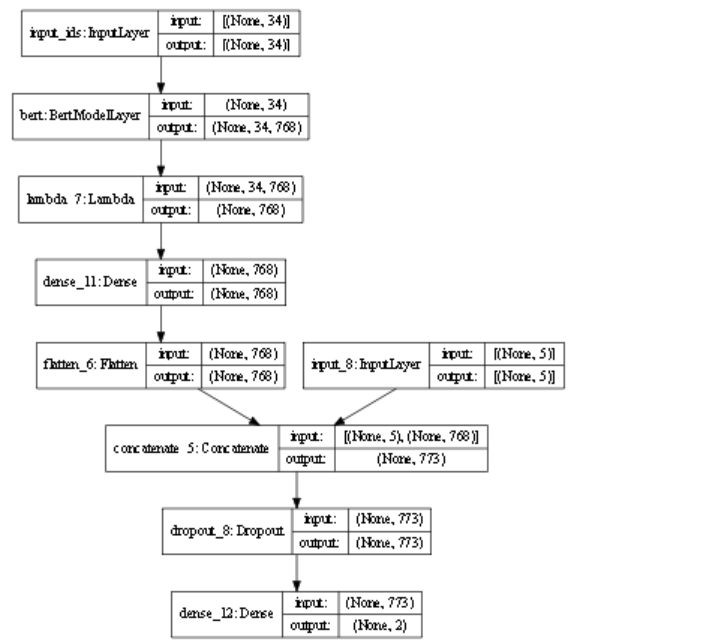}
\caption{Architecture of model for experiment~V}\label{fig:bertChart}
\end{figure}

\begin{figure}[!htb]\centering
\includegraphics[width = 0.9\linewidth]{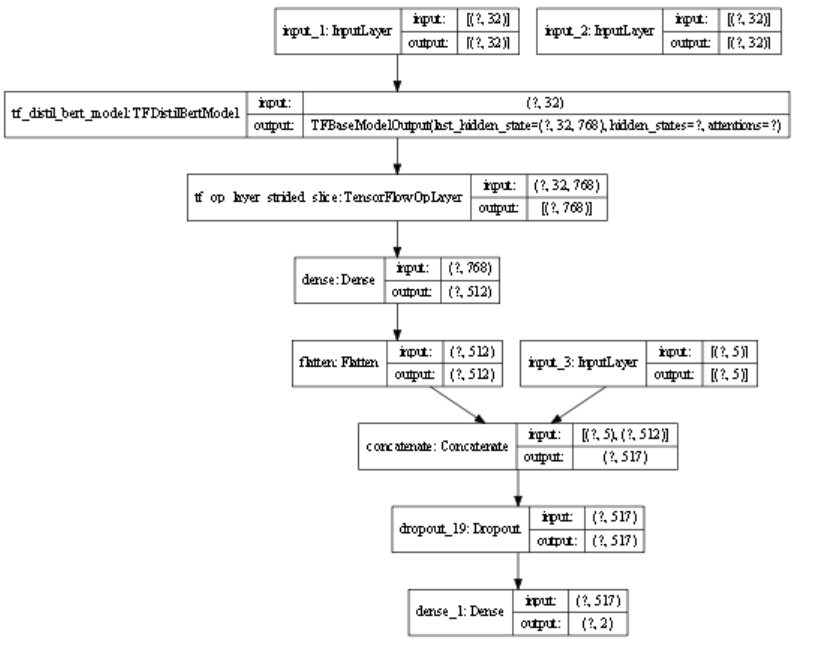}
\caption{Architecture of model for experiment~VI}\label{fig:distilchart}
\end{figure}

\end{document}